# Self-Taught Object Localization with Deep Networks

Loris Bazzani[1]　　Alessandro Bergamo[1]　　Dragomir Anguelov[2]　　Lorenzo Torresani[1]
[1]Department of Computer Science, Dartmouth College　　[2]Google Inc.
{loris.bazzani,alessandro.bergamo.gr,lt}@dartmouth.edu　　dragomir@google.com

## Abstract

*This paper introduces* self-taught object localization, *a novel approach that leverages deep convolutional networks trained for whole-image recognition to localize objects in images without additional human supervision,* i.e., *without using any ground-truth bounding boxes for training. The key idea is to analyze the change in the recognition scores when artificially masking out different regions of the image. The masking out of a region that includes the object typically causes a significant drop in recognition score. This idea is embedded into an agglomerative clustering technique that generates self-taught localization hypotheses. Our object localization scheme outperforms existing proposal methods in both precision and recall for small number of subwindow proposals (e.g., on ILSVRC-2012 it produces a relative gain of 23.4% over the state-of-the-art for top-1 hypothesis). Furthermore, our experiments show that the annotations automatically-generated by our method can be used to train object detectors yielding recognition results remarkably close to those obtained by training on manually-annotated bounding boxes.*

## 1. Introduction

Object recognition, one of the fundamental open challenges of computer vision, can be defined in two subtly different forms: 1) *whole-image classification* [17], where the goal is to categorize a holistic representation of the image, and 2) *detection* [30], which instead aims at decomposing the image into a set of regions or subwindows individually tested for the presence of the target object. Object detection provides several benefits over holistic classification, including the ability to localize objects in the image, as well as robustness to irrelevant visual elements, such as uninformative background, clutter or the presence of other objects. However, while whole-image classifiers can be trained with image examples labeled merely with class information (e.g,, "chair" or "pedestrian"), detectors require richer annotations consisting of manual selections specifying the region or the bounding box containing the target objects in each individual image. Unfortunately, such detailed annotations are expensive and time-consuming to acquire. This effectively limits the applicability of detectors to scenarios involving only few categories (*e.g.*, [12]). Furthermore, these manual selections are often rather subjective and noisy, and as such they do not provide optimal ground truth regions for training detectors.

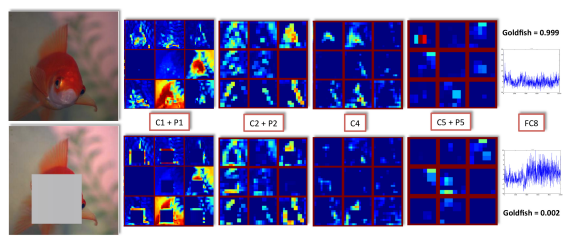

Figure 1: The image (first row), its mask-out version (second row) and the outputs of different layers of the convolutional network [17]. For each layer, a $3 \times 3$ grid of convolutional kernel responses are shown (C=convolution, P=pooling, FC=fully-connected). The final recognition score for the goldfish class is reported at the far right.

Conversely, image class labels are much easier to obtain. This has enabled the creation of huge annotated datasets, such as ImageNet [1], which in turn have spurred dramatic advances in object recognition [17, 27]. Furthermore, recent work has shown that models and features learned using class labels can be effectively leveraged to improve other tasks, such as detection [12], weakly-supervised localization [31], and attribute classification [34]. The transferring of knowledge from pretrained image categorization models to other domains has been so successful that it has become the *de facto standard* for several vision problems.

We propose *self-taught object localization*, an object localization method that leverages a whole-image convolutional network [17] trained on a large collection of class-labeled examples *without* object location information. The key idea is to analyze how the recognition score of the classifier varies as we artificially mask-out regions in the image. When the region containing the object is artificially occluded the whole-image classification score will drop significantly. Figure 1 shows how the partial masking out of the image is propagated through the convolutional network effecting the recognition score. This idea is embedded into a hierarchical clustering technique similar to [30], which merges regions according to their relative drop in recognition score. This produces for each image a set of subwindows that are deemed likely to contain the object.

The proposed method combines bottom-up grouping with top-down (discriminative) information given by the convolutional network. Moreover, we demonstrate that it can be used in scenarios where the object label of the image is not provided by analyzing the top-predicted classes.

The experiments on the ILSVRC-2012 dataset [1] show that our method outperforms the state-of-the-art objectness approaches in terms of recall and precision when considering a small budget of proposals. We obtained a relative increment of 23.4% in terms of top-1 recall with respect to the state of the art. Moreover, a naive combination of our approach with proposal methods optimized for high recall, yields state-of-the-art results for any number of proposals in the range from 1 to $10^4$. We also show that our self-taught localization model trained on the ILSVRC-2012 classes is able to generalize effectively to the different categories of the PASCAL 2007 dataset [8]. Finally, we demonstrate that the subwindows automatically-generated by our approach can be used as positive training examples to learn object detectors without any additional human supervision. Our detection results on 200 classes of ILSVRC-2012 are close to those obtained with the same detection model trained on manually-annotated bounding boxes.

## 2. Related work

Several successful attempts have been made to apply deep networks to object localization and detection problems [12, 27, 29, 7]. In [12], a convolutional network [17] is fine-tuned on ground truth bounding boxes and then applied to classify subwindows generated by selective search [30]. In [27, 7, 29], the network is trained to perform regression directly on the vector-space of bounding boxes. These deep networks have shown promising results compared to standard detection schemes relying on hand-crafted features (*e.g.*, [10, 30]). However, all of these approaches require manually-annotated bounding boxes as training data. In contrast, our method automatically populate the images with bounding boxes likely to contain object, which can then be used for training detectors. We effectively replace the traditional manually-selected bounding boxes with regions automatically estimated from training images annotated only with class labels, which are easy to obtain even for a large number of training images. This framework enables scalable training of object detectors at a much reduced annotation cost. The idea of exploiting image class labels for weakly-supervised localization was also explored in [23]. In contrast, our method does not require to be retrained or finetuned to generate object proposals.

Objectness methods [2, 6, 30, 4, 3, 16, 36, 25, 21, 26, 32, 14] aim at generating bounding boxes that yield high recall for a high number of candidates, *i.e.*, they maximize the probability that each object in the image is covered by at least one subwindow. These methods perform well at testing time and successfully replace the computationally expensive sliding window approach. However, they cannot be used in lieu of ground truth bounding boxes to train a detector because of their low precision caused by the presence of many false positives. Recently, convolutional networks were used for region proposal [18, 24]. However we note that these methods require ground truth bounding boxes or regions during training and thus address a different task compared to ours. Our algorithm can be viewed as a *class-specific* subwindow proposal method which provides precision superior to that of prior methods for low number of candidates. The precision of our approach is high enough that detectors trained on our automatically-generated bounding boxes perform nearly on par with detectors learned from ground-truth annotations. Moreover, during the training of detectors, the class label of each training image is known. Our method exploits this information to generate object-specific proposals which result in better training of detectors compared to generic proposals.

Even though deep networks have shown impressive results, there is still little understanding of what are the critical factors contributing to their outstanding performance. In order to better comprehend deep networks, previous work proposed to visualize the intermediate representations [33, 28], give semantic interpretation of individual units [19], study the emergence of detectors [35] or fool them with artificial images [22]. Liu and Wang [20] have also analyzed what a classifier has learned but for the specific case of bag of features and SVM. Instead, we study the effects of selectively masking out the input of deep networks, which can provide new insights on what the network has learned and how this can be exploited for object localization.

The idea of masking out the input of deep networks has been explored in [33, 13, 5]. [33] investigates the correlation between occlusion of image regions and classification score for the purpose of visualizing the learned features. Although [33] did not provide quantitative results on the task of localization, in our experiments we adapted their occlusion-box strategy to perform object localization but we found that this yields much poorer results compared to our approach (see Section 4 for details). The methods in [13, 5] mask out the background to better focus on foreground features. Our idea is complementary, since we exploit the foreground mask-out mechanism not simply as a feature analysis tool but also as an effective procedure to perform object localization. The method proposed in [28] computes a *class-specific* saliency maps by identifying the pixels that are most useful to predict the classification score of a deep network. Instead, our approach provides state-of-the-art results even when used in a weakly-labeled setup.

## 3. Self-Taught object localization

The aim of Self-Taught Localization (in brief STL) is to generate bounding boxes that are very likely to contain objects. The proposed approach relies on the idea of masking out regions of an image provided as input to a deep network. The drop in recognition score caused by the masking out is embedded into an agglomerative clustering method which merges regions for object localization.

**Input mask-out.** Let us assume to have a deep network $f : \mathbb{R}^N \mapsto \mathbb{R}^C$ that maps an image $\mathbf{x} \in \mathbb{R}^N$ of $N$ pixels to a confidence vector $\mathbf{y} \in \mathbb{R}^C$ of $C$ classes. The confidence vector is defined as $\mathbf{y} = [y_1, y_2, \ldots, y_C]^T$, where $y_i$ corresponds to the classification score of the $i$-th class. We propose to mask out the input image $\mathbf{x}$ by replacing the pixel values in a given rectangular region of the image $\mathbf{b} = [b_x, b_y, w, h] \in \mathbb{N}^4$ with the 3-dimensional vector $\mathbf{g}$ (one dimension for each image channel), where $b_x$ and $b_y$ are the $x$ and $y$ coordinates and $w$ and $h$ are the width and height, respectively. The masking vector $\mathbf{g}$ is learned from a training set as the mean value of the individual image channels. We denote the function that masks out the image $\mathbf{x}$ given the region $\mathbf{b}$ using the vector $\mathbf{g}$ as $h_g : \mathbb{R}^N \times \mathbb{N}^4 \mapsto \mathbb{R}^N$. Please note that the output of the function is again an image (see Figure 1). The bounding boxes $\mathbf{b}$ are automatically generated by our agglomerative clustering method (see details below).

We define the *variation* in classification score of the image $\mathbf{x}$ subject to the masking out of a bounding box $\mathbf{b}$ as the output value of function $\delta_f : \mathbb{R}^N \times \mathbb{N}^4 \mapsto \mathbb{R}^C$ given by

$$\delta_f(\mathbf{x}, \mathbf{b}) = \max(f(\mathbf{x}) - f(h_g(\mathbf{x}, \mathbf{b})), \mathbf{0}) \quad (1)$$

where the max and the difference operators are applied component-wise. This function compares the classification scores of the original image to those of the masked-out image. Intuitively, if the difference for the $c$-th class is large, the masked-out region is very discriminative for that class. Thus the region $\mathbf{b}$ is likely to contain the object of class $c$.

We use the function $\delta_f$ to define two variants of *drop* in classification score, depending on the availability of class label information for the image. When the ground truth class label $c$ of $\mathbf{x}$ is provided, we define the drop function $d_{\text{CL}} : \mathbb{R}^N \times \mathbb{N}^4 \mapsto \mathbb{R}$ as

$$d_{\text{CL}}(\mathbf{x}, \mathbf{b}) = \delta_f(\mathbf{x}, \mathbf{b})^T \mathbb{I}_c, \quad (2)$$

where $\mathbb{I}_c \in \mathbb{N}^C$ is an indicator vector with 1 at the $c$-th position and zeros elsewhere. This drop function enables us to generate *class-specific* proposals in order to populate a training set with bounding boxes likely to contain instances of class $c$. We denote the method which uses $d_{\text{CL}}$ as STL$_{\text{CL}}$.

If the class information is not available, *e.g.* when testing a detector, we use the top-$C_I$ classes predicted by the whole-image classifier $f$ to define $d_{\text{WL}} : \mathbb{R}^N \times \mathbb{N}^4 \mapsto \mathbb{R}$ as

$$d_{\text{WL}}(\mathbf{x}, \mathbf{b}) = \delta_f(\mathbf{x}, \mathbf{b})^T \mathbb{I}_{\text{top-}C_I}, \quad (3)$$

where $\mathbb{I}_{\text{top-}C_I} \in \mathbb{N}^C$ is an indicator vector with ones at the top-$C_I$ predictions for the image $\mathbf{x}$ and zeros elsewhere. Since the function relies on estimated class labels, the setup is called STL$_{\text{WL}}$ where WL stays for *weakly labeled*. In our experiments, we used the top-5 predictions of the deep network $f$ applied to the whole image by leveraging the high recognition accuracy of [17] (the probability of getting the correct class in the top-5 is 82%).

As deep convolutional network $f$ we adopt the model introduced in [17] which has been proven to be very effective for image classification. Since the network is applied to mean-centered data, replacing a region of the image with the learned mean RGB value is effectively equivalent to zeroing out that section of the network input as well as the corresponding units in the hidden convolutional layers (see Figure 1). We want to point out that our masking-out approach is general and it can be applied to any other classifier that operates on raw pixels.

**Agglomerative clustering.** The proposed agglomerative clustering approach is similar to that described in [30]. Specifically, as in [30], we also employ the segmentation method proposed in [11] to generate the initial set of $K$ rectangular[1] regions $\{\mathbf{b}_1, \mathbf{b}_2, \ldots \mathbf{b}_K\}$. Then the goal is to fuse regions (bottom-up) and generate windows that are likely to contain objects (top-down). The main difference with respect to [30] is in the choice of the similarity used to fuse regions.

We propose an iterative method that greedily compares the available regions, and at each iteration merges the two regions that maximize the similarity function discussed below. This procedure terminates when only one region (covering the whole image) is left. The set of generated subwindows are then sorted according to the drop in classification (Eq. 2 for STL$_{\text{CL}}$ and Eq. 3 for STL$_{\text{WL}}$). We also perform non-maximum suppression of the subwindows with overlap greater than 50%.

We define the similarity between regions using four terms capturing the intuitions expressed below. Two bounding boxes are likely to contain parts of the same object if

1. they cause similar large drops in classification score:

$$s_{\text{drop}}(\mathbf{x}, \mathbf{b}_i, \mathbf{b}_j) = 1 - |d_m(\mathbf{x}, \mathbf{b}_i) - d_m(\mathbf{x}, \mathbf{b}_j)| \cdot \max(1 - d_m(\mathbf{x}, \mathbf{b}_i), 1 - d_m(\mathbf{x}, \mathbf{b}_j))$$

---
[1]Note that we mask out the bounding boxes enclosing the segments rather than the segments themselves. We found experimentally that if we mask out the segments, the shape information of the segment is preserved and used by the network to perform recognition, thus causing less substantial drops in classification.

2. they are similar in appearance:

$$s_{\text{app}}(\mathbf{x}, \mathbf{b}_i, \mathbf{b}_j) = z(\phi(\mathbf{x}, \mathbf{b}_i), \phi(\mathbf{x}, \mathbf{b}_j))$$

3. they cover the image as much as possible, encouraging small windows to merge early (as in [30]):

$$s_{\text{size}}(\mathbf{x}, \mathbf{b}_i, \mathbf{b}_j) = 1 - \frac{\text{size}(\mathbf{b}_i) + \text{size}(\mathbf{b}_j)}{\text{size}(\mathbf{x})}$$

4. they are spatially near each other (as in [30]):

$$s_{\text{fill}}(\mathbf{x}, \mathbf{b}_i, \mathbf{b}_j) = 1 - \frac{\text{size}(\mathbf{b}_i \cup \mathbf{b}_j) - \text{size}(\mathbf{b}_i) - \text{size}(\mathbf{b}_j)}{\text{size}(\mathbf{x})}$$

where the index $m \in \{\text{CL}, \text{WL}\}$ in the first term selects $\text{STL}_{\text{CL}}$ or $\text{STL}_{\text{WL}}$ presented in the previous subsection, $z(\cdot, \cdot)$ is the histogram intersection similarity between the network features extracted by $\phi(\cdot, \cdot)$ (see Sec. 4 for details), $\mathbf{b}_i \cup \mathbf{b}_j$ is the bounding box that contains $\mathbf{b}_i$ and $\mathbf{b}_j$. The overall similarity score $s$ is defined as a convex combination of the terms above:

$$s(\mathbf{b}_i, \mathbf{b}_j, \mathbf{x}) = \sum_{l \in \mathcal{L}} \alpha_l \, s_l(\mathbf{b}_i, \mathbf{b}_j, \mathbf{x}), \quad (4)$$

where $\mathcal{L} = \{\text{drop}, \text{app}, \text{size}, \text{fill}\}$ and the $\alpha$s are set to be uniform weights in our experiments. We empirically found that removing $s_{\text{drop}}$ from Eq. 4 will cause a drop of $8\%$ and $10\%$ in terms of precision and recall, respectively.

Figure 2 illustrates the intuition behind the similarity measure encoded by $s_{\text{drop}}$. This similarity is large if the two regions exhibit similar classification drops when occluded (corresponding to points on the diagonal of the $xy$-plane in the 3D plot) and it is especially large when the drop in score is substantial (points close to $(1,1)$ in the plot). The term $s_{\text{app}}$ encourages aggregation of regions similar in appearance, while $s_{\text{size}}$ and $s_{\text{fill}}$ borrowed from [30] favor early merging of small regions and regions that are near each other, respectively.

There are many advantages of the proposed similarity with respect to [30]. First, it does not rely on the hand-engineered features used in [30], but instead it leverages the features learned by the deep network. Moreover, our similarity exploits the discriminant power of the deep convolutional network enabling our method to generate class-specific window proposals. Even when used in the weakly-labeled regime of Eq. 3 it will tend to generate subwindows that are most informative for recognition (since their occlusion causes large classification drops). Thus, our approach can be viewed as a hybrid scheme combining bottom-up cues (size, appearance) with top-down information (object-class recognition), unlike [30] where the merging of regions is driven by a pure bottom-up procedure.

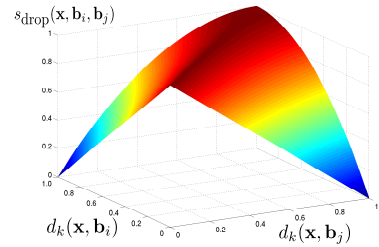

Figure 2: Similarity score $s_{\text{drop}}(\cdot)$ as a function of the drops in classification $d_k(\mathbf{x}, \mathbf{b}_i)$ and $d_k(\mathbf{x}, \mathbf{b}_j)$.

## 4. Experiments

In this section we present comparative results of our approach with state-of-the-art methods on the task of object subwindow proposal. We also show that $\text{STL}_{\text{CL}}$ can be used to generate annotations for training object detectors.

**Implementation details.** In our experiments, we used the convolutional network software *Caffe* [15] with the model trained on ILSVRC-2012 provided by the authors. Inspired by [12], the descriptor ($\phi$) used in the term $s_{\text{app}}$ of STL is the vector from the last fully-connected layer (before the soft-max) of the network.

**Datasets.** Our experiments were carried out on two challenging benchmarks: ILSVRC-2012-LOC [1] and PASCAL-VOC-2007 [8]. ILSVRC-2012-LOC is a large-scale benchmark for object localization containing 1000 categories. The training set contains 544546 images with 619207 annotated bounding boxes. The validation set contains 50000 images for a total of 76750 annotated bounding boxes. PASCAL-VOC-2007 contains 20 categories, for a total of 9963 images divided into training, validation and testing splits. Each image contains multiple objects belonging to different categories at different positions and scales, for a total of 24640 ground truth bounding boxes.

**Object proposal.** Given a test image, the goal is to generate the best set of bounding boxes that enclose the objects of interest with high probability. A true positive is a proposed bounding box whose intersection over union with the ground truth is at least $50\%$ [8]. The performance is then measured in terms of the mean of the average recall and precision *per class* [30] as done in the PASCAL benchmark [8].

We compared STL to recent state-of-the-art proposal methods: SELSEARCH [30] (*fast* version), BING [4] (*MAXBGR* version), EDGEBOXES [36] and MCG [3]. We note that these prior proposal methods do not make use of image class label when proposing subwindows. Thus, the supervised version of STL ($\text{STL}_{\text{CL}}$) is in a sense given an unfair advantage over them as it can generate class-specific proposals consistent with the ground-truth label. However, we will demonstrate that our weakly-labeled $\text{STL}_{\text{WL}}$ provides results nearly equivalent to $\text{STL}_{\text{CL}}$.

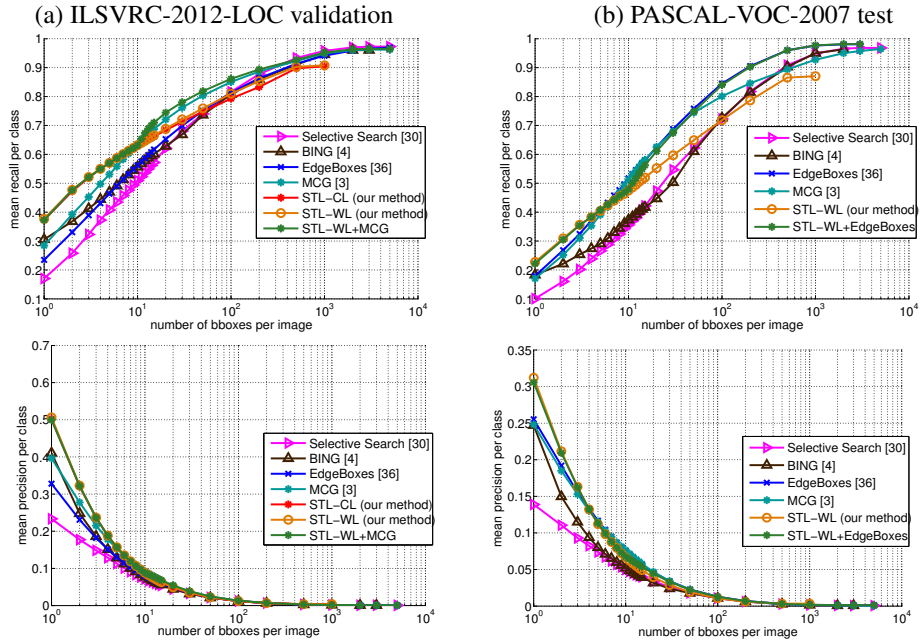

Figure 3: Comparison of different bounding-box proposal methods. The first and second row report the mean recall per class and the mean precision per class, respectively, as a function of the number of proposed subwindows. The columns show two datasets: ILSVRC-2012-LOC (validation) and PASCAL-VOC-2007 (test).

Figure 3(a) reports the results in terms of recall (first row) and precision (second row) on the validation set of ILSVRC-2012-LOC. Note that this dataset is disjoint from the set used to train the convolutional network $f$. Figure 3(a) shows that our method outperforms all the other methods for the first 10 proposed subwindows. We obtained a relative improvement in the top-1 recall of $23.4\%$ over BING, which is the best method for the top-1 case. Figure 3(a) also shows that the performance difference between using the class label of the image ($STL_{CL}$) and not using it ($STL_{WL}$) is negligible. This indicates that STL works equally well even when the class label is not given.

SELSEARCH, BING, EDGEBOXES and MCG were designed to obtain high recall when using a large number of proposals, which is a desirable property at testing time. However it yields precision not sufficiently high to train a detector as shown in Figures 3(a), second row. In contrast, STL is by far the best method in term of both precision and recall for a small number of proposals. In order to capture the diversity between these methods, we carried out an experiment where the top-10 proposals of $STL_{WL}$ are merged with the ones of the best performing method for large number of proposals, that is MCG. This experiment is reported in Figures 3(a) denoted as $STL_{WL}$+MCG. This result demonstrates that we can obtain the best performance on the range 1-10 (where $STL_{WL}$ stands out) as well as competitive results on the range $11\text{-}10^4$ (where MCG is the best) in terms of both recall and precision.

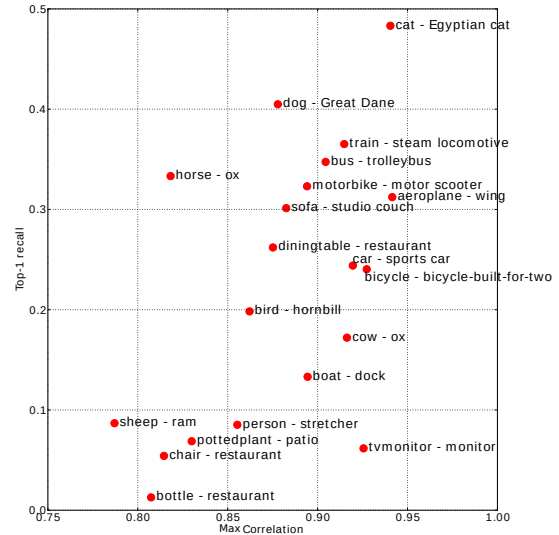

Figure 4: Cross-dataset generalization. Each dot represents a PASCAL class: the $x$ coordinate is its max correlation value against ILSVRC categories, while the $y$ coordinate shows the top-1 recall achieved by STL on that PASCAL category. The text for each dot lists the PASCAL class and its most correlated category in ILSVRC.

We also tested two simple baselines: the sliding window and the sliding occlusion box. In the sliding window approach, a set of rectangles of different sizes is slid over the image and at each position we compute the confidence score as the sum of the classification scores of the top-5 classes predicted by the convolutional network. In the sliding occlusion box approach, occlusion boxes of different scale slide over a fixed grid of the image and uses the corresponding drops in classification score. This is reminiscent of the approach described in [33]. Both approaches perform poorly in terms of recall: $-46\%$ and $-18\%$ in absolute value for the top-1 proposal on ILSVRC-2012-LOC with respect to STL. This shows the need of objectness methods in contrast with sliding window strategies and more importantly the value of the proposed agglomerative clustering.

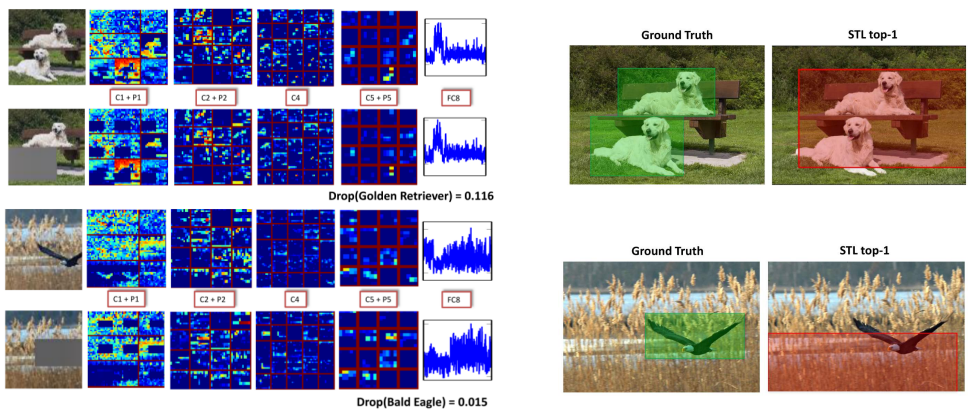

Figure 5: Two cases where the classification score does not drop after masking-out the object. In the dog picture this is due to the presence of multiple objects of the same class while in the bald eagle photo this is caused by background statistically correlated with the object. On the left, we show the original image, a mask-out version of the image and the respective outputs of different layers of the deep convolutional network [17]. On the right, we report the top-1 bounding box generated by STL.

Next, we analyze the ability of the proposed method to generalize to unseen datasets using the PASCAL-VOC-2007 benchmark [8]. The images of this set have very different statistics than those in ILSVRC-2012-LOC, as each image can contain multiple objects belonging to different categories. Unlike BING and EDGEBOXES, STL was not trained or fine-tuned using the images from this dataset. Nevertheless, as shown in Figure 3(b), our method is able to generalize well to this new scenario especially in terms of mean precision (second row). This result shows that STL$_{WL}$ performs well on arbitrary classes, as the categories of PASCAL-VOC-2007 do not exactly correspond to classes in ILSVRC-2012-LOC. Similarly to the experiment performed on ILSVRC-2012-LOC, we carried out an experiment by merging STL$_{WL}$ with the best performing method for PASCAL-VOC-2007, that is EDGEBOXES. The green curve in Figure 3(b) corresponds to this experiments. Also in this case, the combination between the two methods produces the best results highlighting the complementarity of STL$_{WL}$ and other methods.

In order to understand whether the good performance on the PASCAL-VOC-2007 classes is due to these categories being approximately represented in ILSVRC-2012-LOC, we analyzed the correlation between the convolutional network descriptors ($\phi$) of PASCAL and ILSVRC. For each class in both datasets, we computed its average feature descriptor (prototype). Then we computed the cross-dataset correlation between PASCAL prototypes and ILSVRC prototypes using the cosine similarity[2]. For each PASCAL prototype, we recorded the "max correlation" value against the ILSVRC prototypes. Figure 4 plots for each individual PASCAL class this max correlation value ($x$ axis) against the top-1 recall ($y$ axis) achieved by our method on that category. We can observe that the classes with high max correlation have also high recall (*e.g.*, cat - Egyptian cat). On the other hand, the classes that are poorly represented in ILSVRC (i.e., low max correlation) tend to have low recall (*e.g.*, bottle - restaurant). This demonstrates that STL leverages the network trained on 1000 classes to effectively transfer localization to "novel" categories but its performance is generally higher when the novel classes are not too distant from those seen in ILSVRC.

Finally, we analyze qualitatively some cases where STL fails in order to understand better the proposed method. In Figure 5, it is interesting to note how the mask-out operation propagates through the intermediate convolutional layers of the network (*i.e.* rectangular dark blue box in the feature maps), however some feature maps are still very active. The first example (the golden retriever) fails because of the presence of identical instances of the same class. Masking out one of the two dogs causes a small drop since the network is still activated from the other dog instance. In the second case (the bald eagle), the drop is small because the convolutional network most likely uses contextual information (for example the color distribution of the background) that has learned as correlating to the eagle category during training.

**Generating annotations for training detectors.** In this section, we show that the bounding boxes generated by STL$_{CL}$ (Sec. 3) can be exploited as annotations when train-ing object detectors, thus eliminating the need for ground truth annotations. To this end, we use a subset of 200 randomly selected classes from ILSVRC-2012-LOC (which we denote as ILSVRC-2012-LOC-200[3]) as this allowed us to perform faster training, thus enabling a more comprehensive study of the different methods on the detection task.

---
[2]Defined as one minus the cosine distance.

| Positive Boxes + Negative/Test Boxes | BING [4] + BING | EDGEBOXES [36] + EDGEBOXES | MCG [3] + MCG | SELSEARCH [30] + SELSEARCH | STL$_{CL}$ (our) + STL$_{WL}$ (our) | GROUNDTRUTH + SELSEARCH |
|---|---|---|---|---|---|---|
| mAP | 13.78 | 17.03 | 17.94 | 18.31 | **19.60** | 25.40 |

Table 1: The first row reports which method is used to generate the bounding boxes of the positive set, and those of the negative and test set. The second row contains the mean Average Precision (%) calculated as the mean across all 200 classes for ILSVRC-2012-LOC-200.

| Positive Boxes + Negative/Test Boxes | BING [4] + SELSEARCH | SELSEARCH [30] + SELSEARCH | STL$_{CL}$ (our method) + SELSEARCH | GROUNDTRUTH + SELSEARCH |
|---|---|---|---|---|
| mAP (all classes) | 19.55 | 18.31 | **20.43** | 25.40 |
| best classes | leopard = 56.83<br>giant panda = 50.73<br>koala = 49.50<br>car mirror = 48.25<br>orangutan = 46.76<br>pickup = 45.09<br>admiral = 44.69<br>frilled lizard = 44.57<br>entertainment center = 43.04<br>teapot = 42.67 | leopard = 59.29<br>car mirror = 50.86<br>koala = 49.23<br>admiral = 44.96<br>giant panda = 44.19<br>crib = 41.21<br>bullfrog = 41.00<br>maze = 40.67<br>orangutan = 40.66<br>whiskey jug = 38.27 | leopard = 62.86<br>teapot = 57.26<br>giant panda = 54.96<br>car mirror = 51.69<br>Crock Pot = 50.56<br>koala = 50.15<br>police van = 48.37<br>admiral = 46.24<br>necklace = 46.05<br>pickup = 45.75 | leopard = 65.28<br>Crock Pot = 62.60<br>teapot = 59.12<br>admiral = 58.55<br>car mirror = 58.28<br>koala = 55.63<br>cabbage butterfly = 54.73<br>frilled lizard = 52.10<br>police van = 51.86<br>giant panda = 51.68 |
| worst classes | *flute = 0.21*<br>*punching bag = 0.21*<br>*swimming trunks = 0.16*<br>*pole = 0.03*<br>*basketball = 0.01* | *punching bag = 0.12*<br>*hair spray = 0.03*<br>*basketball = 0.01*<br>*pole = 0.01*<br>*nail = 0.01* | *croquet ball = 0.15*<br>*punching bag = 0.13*<br>*basketball = 0.11*<br>*pole = 0.10*<br>*nail = 0.04* | *punching bag = 0.63*<br>*hair spray = 0.54*<br>*screwdriver = 0.41*<br>*nail = 0.10*<br>*pole = 0.05* |

Table 2: Each column contains the best classes (blue) and the *worst classes* (red) for the detectors trained using the annotation method listed at the top, along with the Average Precision (%). All methods were trained on ILSVRC-2012-LOC-200.

200 detectors were trained (one for each class in ILSVRC-2012-LOC-200), using for each a training set of 50 positive images and 4975 negative images (obtained by sampling 25 examples from each negative class). The test set is composed by 10000 images of the ILSVRC-2012-LOC-200 validation set. As detection model, we use the RCNN detector of [12], with the difference that we train it with the simpler negative mining procedure described in [30]. However, while [30, 12] exploited manually-annotated bounding boxes as positive examples, in our training procedure we replace the ground truth regions with the top-$\hat{K}$ bounding boxes produced by the class-specific STL$_{CL}$ on training images of class $c$, *i.e.*, we use the class label information for localization of the positive regions.

The negative set is built using the bounding boxes that overlap less than 30% with any STL$_{CL}$ subwindows from the positive images, and one randomly-chosen bounding box from each negative image. At each iteration, a linear SVM [9] is trained by automatically choosing the hyper-parameter with a 5-fold cross-validation that maximizes the average precision. The negative set is augmented for the next training iteration by adding for each negative image the bounding box with the highest positive score. At testing time, each detector is tested on the generated subwindows of a given image, the detection scores are sorted and then pruned via non-maximum suppression: we remove a subwindow if it overlaps for more than 70% with a subwindow that has higher score.

We experimented with different combinations of proposal methods for the positive and the negative bounding boxes. For each combination, at test time on each input image we used the same proposal method that was applied to generate the negative boxes during training. For all combinations we use the top-3 candidates as positive bounding boxes to obtain a good recall/precision trade-off based on the results of Figure 3.

Table 1 shows the results in terms of mean average precision (mAP) across *all* the 200 classes for each method computed according to the PASCAL VOC criterion [8]. The first row reports the method used to generate the positive training boxes, the second row indicated the method for the negative and test boxes. Our approach is STL$_{CL}$+STL$_{WL}$ (sixth column of Table 1), and it involves using our proposal method based on class labels (since when training a detector they are always available) to generate the positive boxes and our unsupervised approach to produce the negative boxes as well as the proposals on the test images. We compare this approach to BING, EDGEBOXES, MCG and SELSEARCH, where each of these methods was used to generate both the positive boxes as well as the negative and testing boxes of the detector (second to fifth columns of Table 1). We also compared our method to the fully-supervised approach based on manually-annotated positive boxes as proposed in [30] (named GROUNDTRUTH+SELSEARCH in Table 1).

---
[3]To enable future comparisons with our results, we will make publicly available the list of 200 classes.

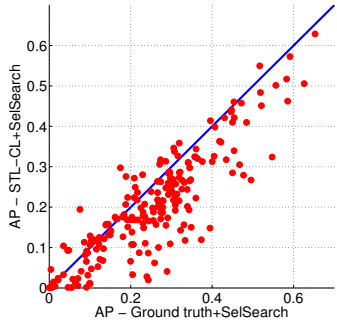

Figure 6: Average precision (AP) on the individual 200 classes obtained with the fully-supervised approach GROUNDTRUTH+SELSEARCH ($x$-axis) and our method STL$_{CL}$+SELSEARCH ($y$-axis). Each point represents the AP of these two methods on one particular class.

We notice that STL$_{CL}$+STL$_{WL}$ outperforms all the other methods, yielding a relative improvement of 42.2% over the worse method (BING) and 7% over the best competitor (SELSEARCH). This suggests that STL$_{CL}$ generates reliable bounding boxes for the positive set.

An interesting observation that can be drawn from Figure 3(a) is that while STL$_{CL}$ produces the highest precision for a small number of subwindows (and therefore it is the preferred method to generate the positive boxes), other methods yield higher recall for large numbers of proposals. This suggests that using, for example, SELSEARCH for the negative and test images can be advantageous. Based on this observation, we performed an experiment where we tested "the best of the two worlds", i.e., using STL$_{CL}$ to generate the positive set and SELSEARCH for the negative and test set (Table 2, fourth column). We also tested combination of other proposal methods for positive subwindows with SELSEARCH for negative/test subwindows and reported the results on Table 2 (second and third column). Table 2 (second row) shows that STL$_{CL}$+SELSEARCH outperforms all the other combinations. Moreover, STL$_{CL}$+SELSEARCH shows a relative drop in performance of only 19.6% with respect to the fully supervised method (last column). This is a remarkable result given that it uses only class labels. We also tested STL$_{CL}$+SELSEARCH using the top-1 bounding box obtaining a mAP of 20.93%, which reduces to 17.6% the relative gap with respect to the fully-supervised method.

Table 2 shows also the best-10 and worst-5 classes for each method along with the respective APs. It is interesting to notice that 8 out of the 10 best categories are shared between the detectors trained on the ground truth annotations (last column) and our STL$_{CL}$ (forth column) as opposed to 5 out of 10 of our competitors.

In Figure 6 we report the AP on each individual class for the proposed method STL$_{CL}$+SELSEARCH ($y$-axis) and the fully-supervised approach GROUNDTRUTH+SELSEARCH ($x$-axis). For 41 classes (all points above the diagonal) the proposed method achieves better accuracy than that obtained when using ground truth annotations.

**Analysis of computational costs.** Let $K$ be the number of segments produced by the method of [11]. During initialization, the similarity of Eq. 4 is evaluated for all segment pairs, for a total of $O(K^2)$ times. However, note that only $K$ evaluations of the convolutional network are needed, one for each masked-out segment (Eq. 1).

At the first iteration of the clustering procedure, two of the segments are merged, and there will be $K-1$ remaining segments. Only the similarities involving the newly created segment are updated, which amount to $O(K)$ similarity evaluations, but these can be obtained with a single network evaluation of the image with only the newly merged segment masked-out. In every subsequent iteration, the total number of segments will decrease by one. Thus, in total only $2 \cdot K$ network evaluations are performed over the entire procedure, including those done at initialization.

In practice, the $2 \cdot K$ network evaluations of an image take about 210 seconds on CPU or 20 seconds on GPU for typical values of $K$ using our non-optimized Python code. MCG, EDGEBOXES, SELSEARCH and BING are highly optimized and they take 25, 0.25, 10 and 0.2 seconds per image, respectively. The runtime of STL can be optimized by merging only adjacent segments during agglomerative clustering. Note that most of the computation is done during the initialization of the clustering algorithm (when $K$ mask-out operations are performed). At the same time, these segments are very small and therefore most of the ConvNet features with limited receptive field do not change. We leave the optimization of the STL code as future work.

## 5. Conclusions

This work presents self-taught localization, which leverages the power of convolutional networks trained using image class labels to automatically object proposal subwindows. We showed that STL outperforms the state-of-the-art methods on the task of object localization. We demonstrated that detectors trained on localization hypotheses automatically generated by STL achieve performance nearly comparable to those produced when training on manually selected bounding boxes. In future work we will investigate the possibility of fine-tuning the network as a localizer on the subwindows generated by STL and how to use them in a multiple instance learning framework in order to have more robust object detectors. The code of our method will be made publicly available.

**Acknowledgements.** We thank Haris Baig for helpful discussion. This work was funded in part by Google, NSF award CNS-1205521 and a CompX Faculty Grant from the William H. Neukom 1964 Institute for Computational Science.

# Supplementary Material

## Abstract

*In this supplementary material we provide further evidence that supports the quality of the proposed method. These additional experiments were produced using the same version of the algorithm explained in our paper and include:*

- *Visualization of the mask-out effect in terms of convolutional feature maps and drop in classification (Figures 1 and 2);*

- *Visualization of top-1 bounding boxes generated by STL$_{CL}$ (our method) and SELSEARCH [1] (Table 1).*

*Note: this document is best viewed in color.*

## Visualizing the Mask-out Effect

We show some qualitative examples of the effect of the mask-out operation on images in Figure 1 and 2. Each row reports network input (*i.e.*, the image) and feature maps from each of the 5 convolutional layers of the network (shown as a grid of $F \times F$ feature maps). The first row in each set shows the original images, while the second row shows the effects on the the masked-out image. We also report the value of the *drop* in classification (Eq. 1 in the paper) caused by to the mask-out operation.

In Fig. 1, we can see some cases where the proposed method succeeds, i.e., where masking out the object region causes a significant drop in classification score. It is interesting to visually note how the mask-out operation propagates through the intermediate convolutional layers of the net until reaching the classification output (as evidenced by the drop). The mask-out operation essentially corresponds to zeroing out the feature map values corresponding to pixels in the masked-out region (*e.g.* rectangular dark blue box in *norm1* feature maps).

Fig. 2 shows some hard examples where our localization method is prone to *fail* because the drop in recognition is not high. In Fig. 2(a), masking out one of the two dogs causes a small drop since there is still one dog that can be recognized by the network. Moreover, a small drop may happen also when the convolutional network uses contextual information (for example the color distribution of the background) that has learned as correlating to some specific category during training, *e.g.*, the eagle and the background landscape in Fig. 2(b). Finally, the basketball example in Fig. 2(c) shows that the network is still able to classify the object (the ball) even when the object of interest is masked out. This is due to the frequent co-occurrence in the training set of basketball and basketball player. The network therefore learned the co-occurrence of the two different objects but not the characteristic of the basketball itself. Fortunately, because STL relies on three other terms it can propose good subwindows also in cases where the mask-out term fails.

We show in Table 1 the top-scoring bounding box on a few sample images of the dataset ILSVRC-2012-LOC, using different bounding box proposal methods. In the case of our method (STL$_{CL}$), we show the top bounding box selected according to Eq. 1 in the paper. As already highlighted by the quantitative results, the subwindows produced by STL$_{CL}$ are more accurate than those produced by SELSEARCH. It is also interesting to notice in the last row of the table that multiple similar instances of the same object are often grouped together because STL$_{CL}$ yields the maximum drop in classification when all of them are masked out.

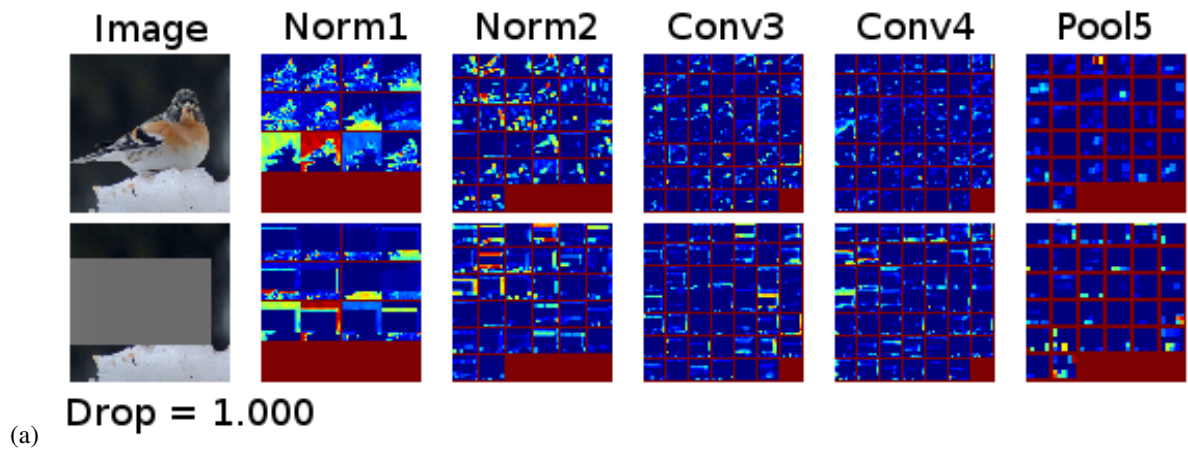

(a)

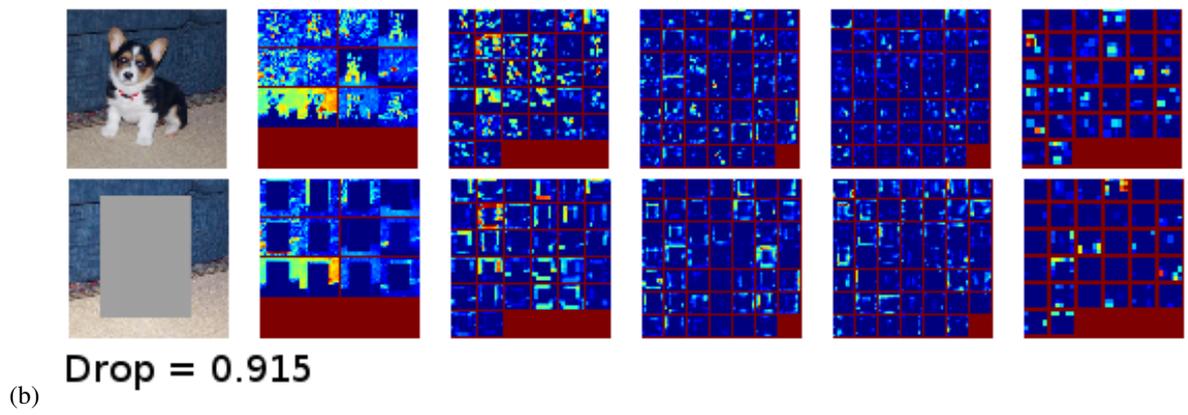

(b)

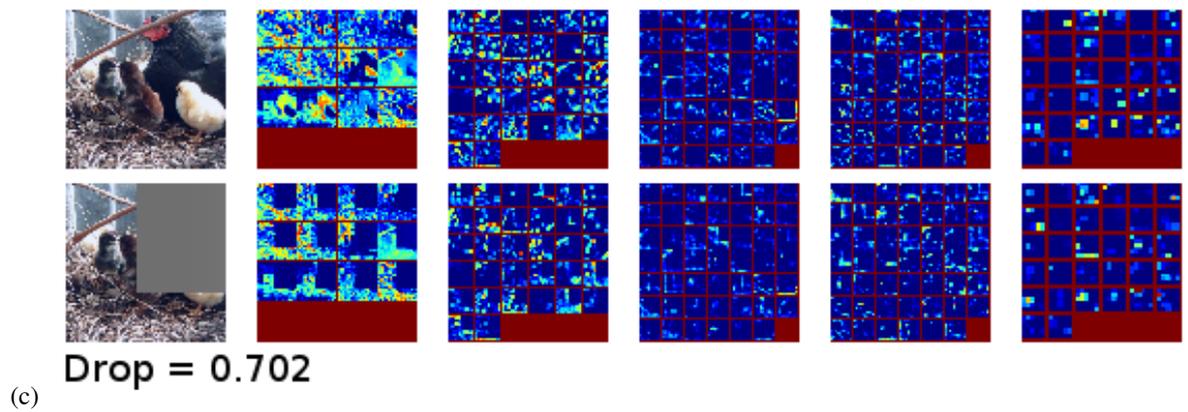

(c)

Figure 1. Successful examples where masking out the object yields large drops in classification score. Each row reports network input (*i.e.*, the image) and the feature maps from each of the 5 convolutional layers of the network (shown as a grid of $F \times F$ feature maps). The first row in each set shows the original image, the second row the masked-out image.

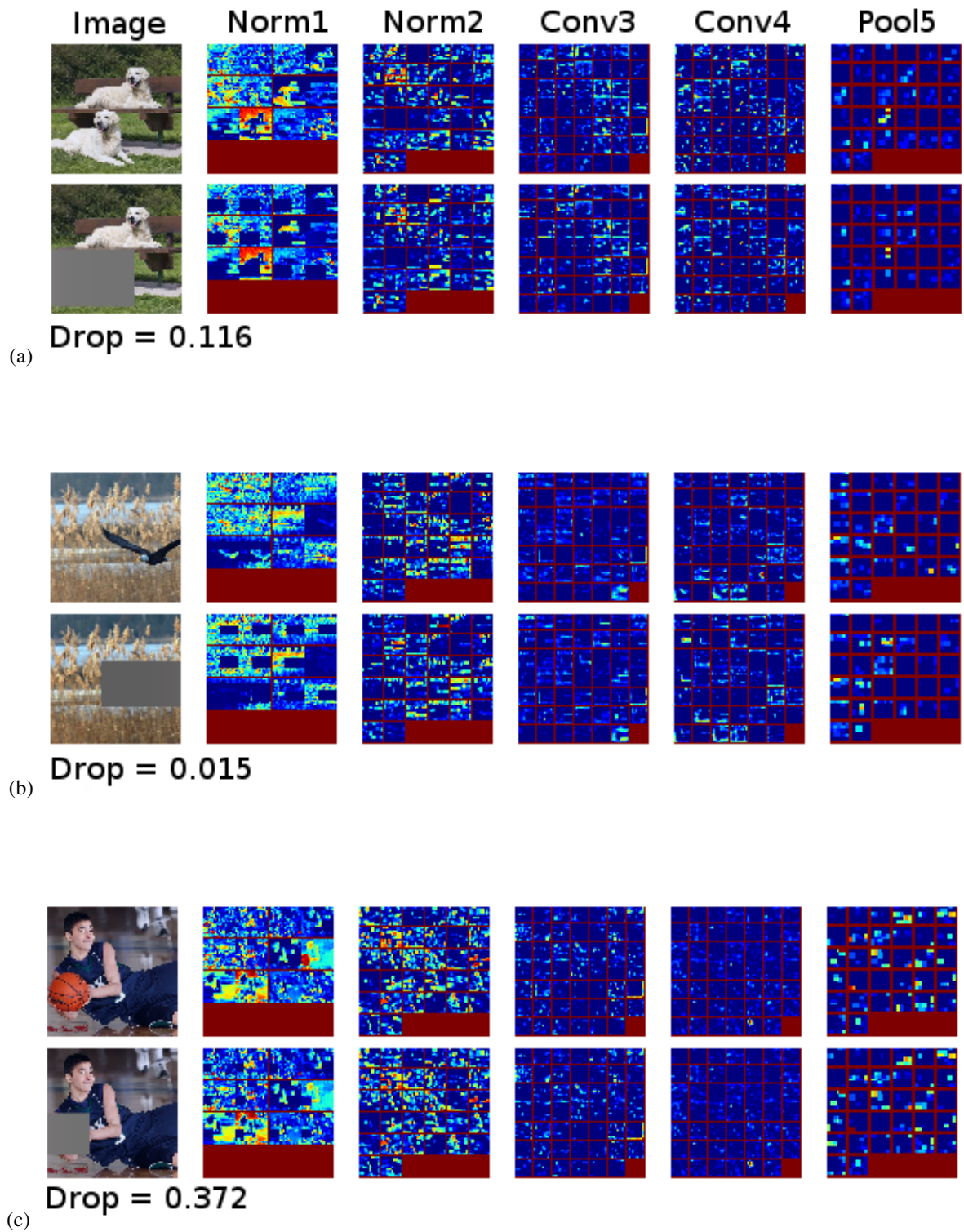

Figure 2. Cases where the masking out of the object fails to significantly drop the classification score due to multiple objects (a) or the presence of context useful to recognize the object, such as in (b) and (c).

| GT | STL$_{CL}$ | SELSEARCH [1] |
|---|---|---|
| 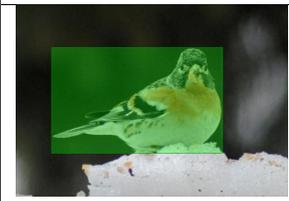 | 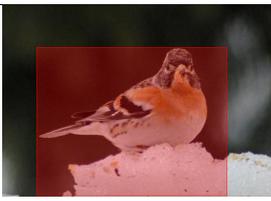 | 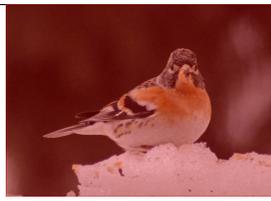 |
| 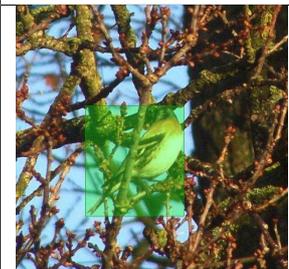 | 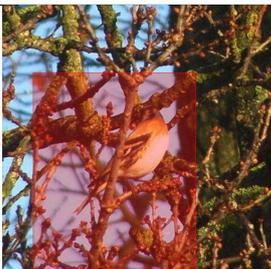 | 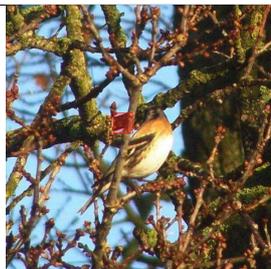 |
| 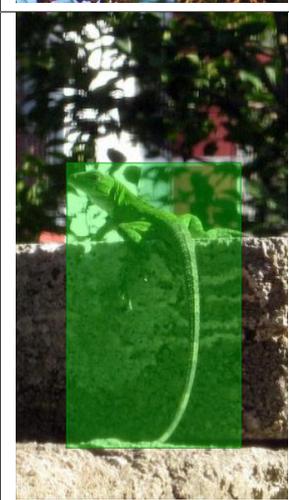 | 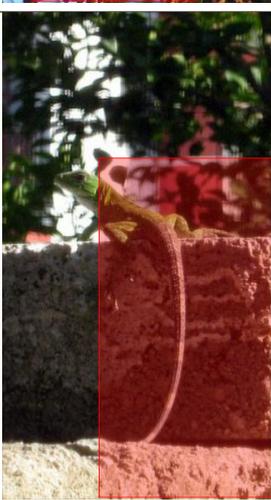 | 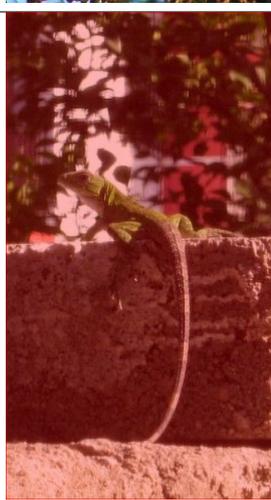 |
| 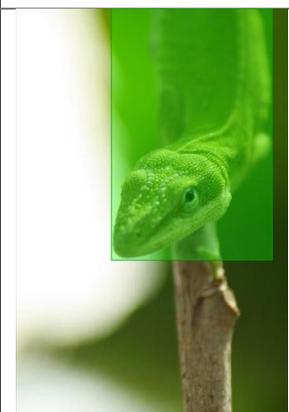 | 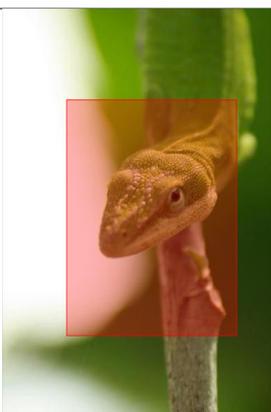 | 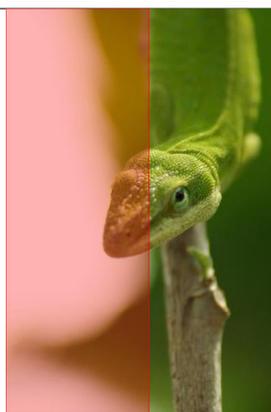 |
| 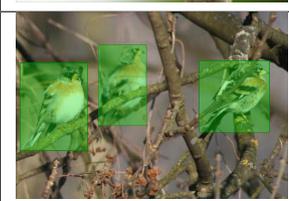 | 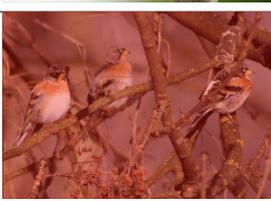 | 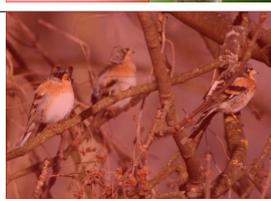 |

Table 1. Top-scoring bounding boxes generated by STL$_{CL}$ and SELSEARCH for a few sample images from the dataset ILSVRC-2012-LOC-200.

.